\documentclass[conference]{IEEEtran}
\usepackage{times}

\usepackage[numbers]{natbib}
\usepackage{multicol}

\usepackage[bookmarks=true]{hyperref}
\usepackage{amsmath,amssymb,amsfonts}
\usepackage{algorithmic}
\usepackage{graphicx}
\usepackage{todonotes}
\usepackage{bm}
\usepackage{pifont}
\usepackage{booktabs}
\usepackage{etoc}

\renewcommand{\vec}[1]{\bm{#1}}

\newcommand{\acro}[0]{NDCF}

\begin{document}

\title{Integrated Object Deformation and Contact Patch Estimation from Visuo-Tactile Feedback}


\author{\authorblockN{Mark Van der Merwe, Youngsun Wi, Dmitry Berenson, Nima Fazeli}
\authorblockA{University of Michigan, Ann Arbor, MI 48109\\
Email: \{markvdm, yswi, dmitryb, nfz\}@umich.edu}}

\maketitle

\begin{abstract}
Reasoning over the interplay between object deformation and force transmission through contact is central to the manipulation of compliant objects. In this paper, we propose Neural Deforming Contact Field (NDCF), a representation that jointly models object deformations and contact patches from visuo-tactile feedback using implicit representations. Representing the object geometry and contact with the environment implicitly allows a single model to predict contact patches of varying complexity. Additionally, learning geometry and contact simultaneously allows us to enforce physical priors, such as ensuring contacts lie on the surface of the object. We propose a neural network architecture to learn a NDCF, and train it using simulated data. We then demonstrate that the learned NDCF transfers directly to the real-world without the need for fine-tuning. We benchmark our proposed approach against a baseline representing geometry and contact patches with point clouds. We find that NDCF performs better on simulated data and in transfer to the real-world. More details and video results can be found at \url{https://www.mmintlab.com/ndcf/}.\footnote{This work was supported in part by Toyota Research Institute, the Office of Naval Research Grant N00014-21-1-2118 and NSF grants IIS-1750489, IIS-2113401, and  IIS-2220876. This material is based upon work supported by the National Science Foundation Graduate Research Fellowship Program under Grant No. 1841052. Any opinions, findings, and conclusions or recommendations expressed in this material are those of the authors and do not necessarily reflect the views of the National Science Foundation.}
\end{abstract}

\IEEEpeerreviewmaketitle

\section{Introduction}

The ability to manipulate elastically deformable objects (e.g., spatulas and sponges) is crucial for many contact-rich manipulation tasks. In order for robots to effectively use these compliant tools, they must reason over the coupled dynamics of object deformation and environmental contact to control the resulting contact interface between the two.
However, there are two key challenges to address. First, the frictional interactions between these objects and their environment is governed by complex non-linear mechanics, making it challenging to model and control their behavior. Second, perception of these objects is challenging due to both self-occlusions and occlusions that occur at the contact location (e.g., when wiping a table with a sponge, the contact is occluded). 



\begin{figure}
    \centering
    \includegraphics[width=\linewidth]{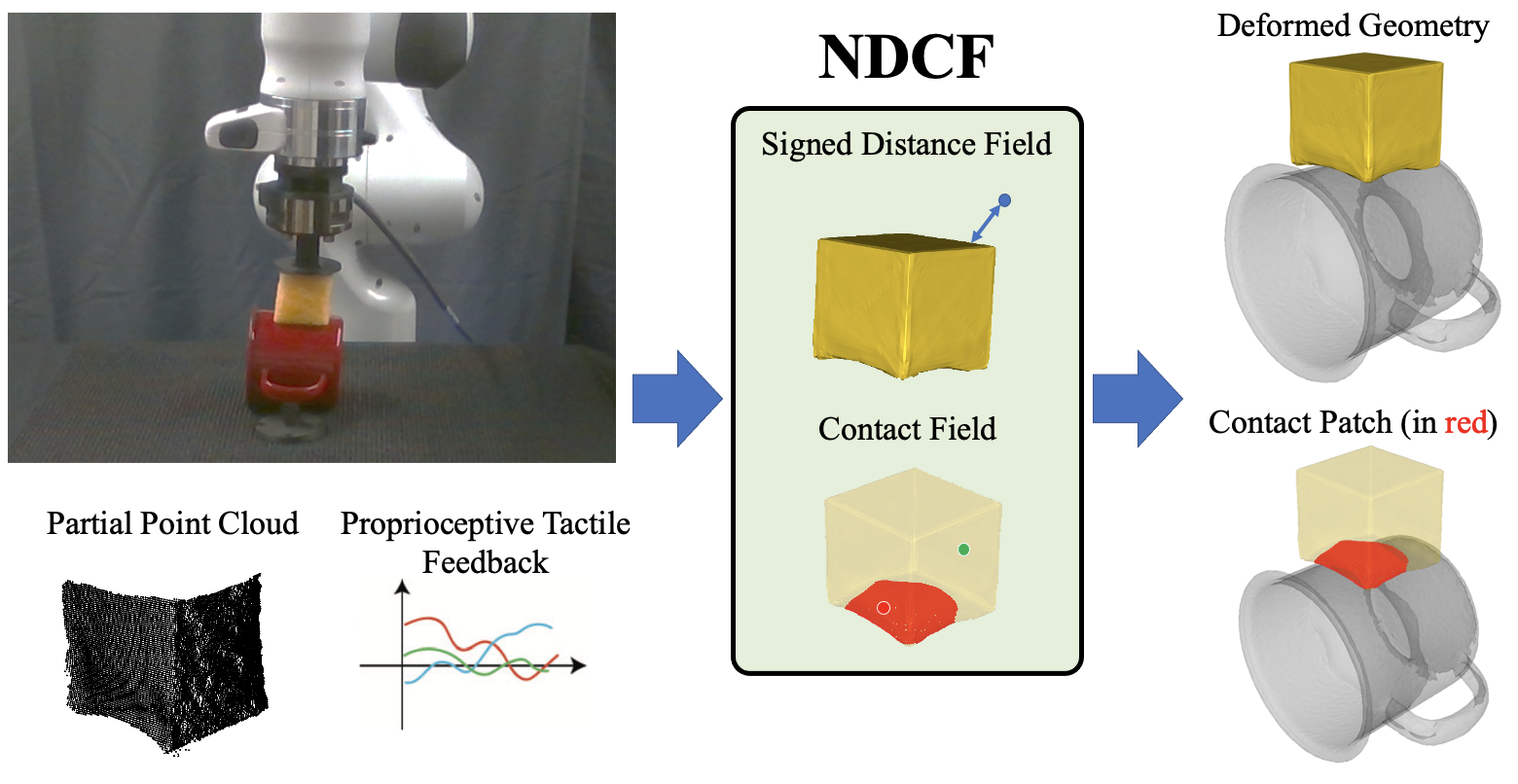}
    \caption{We present Neural Deforming Contact Fields (NDCFs), a method for recovering deformation and contact patch predictions from partial point clouds and tactile feedback (here, robot proprioceptive tactile feedback). 
    We show that while NDCF is trained using simulated data, it is able to transfer to the real-world directly. Here, a robot with a deforming sponge presses into the YCB mug object. NDCF is able to faithfully recover the deformed tool geometry and contact patch between the deformed sponge and rigid mug (shown in red) given real-world sensing from this interaction.}
    \label{fig:intro_figure}
\end{figure}

In this paper, we propose Neural Deforming Contact Fields (\acro) -- a visuo-tactile neural implicit representation that models both the object and contact patch geometries as implicit fields (see Fig.~\ref{fig:intro_figure}). Specifically, our approach maps each point in 3D space to a signed-distance value and a probability of contact. By jointly reasoning over these coupled fields, our representation enables: i) modeling complex object and contact formation geometries; ii) enforcing physical priors such as ensuring that contacts lie on the surface of the object; and iii) estimating contact patches or deformed geometries given partial visuo-tactile observations. We further present a neural network implementation and learning algorithm for \acro. Finally, we evaluate \acro~on simulated and real-world data and benchmark against explicit methods utilizing point cloud representations.

\begin{figure*}
    \centering
    \includegraphics[width=0.99\textwidth]{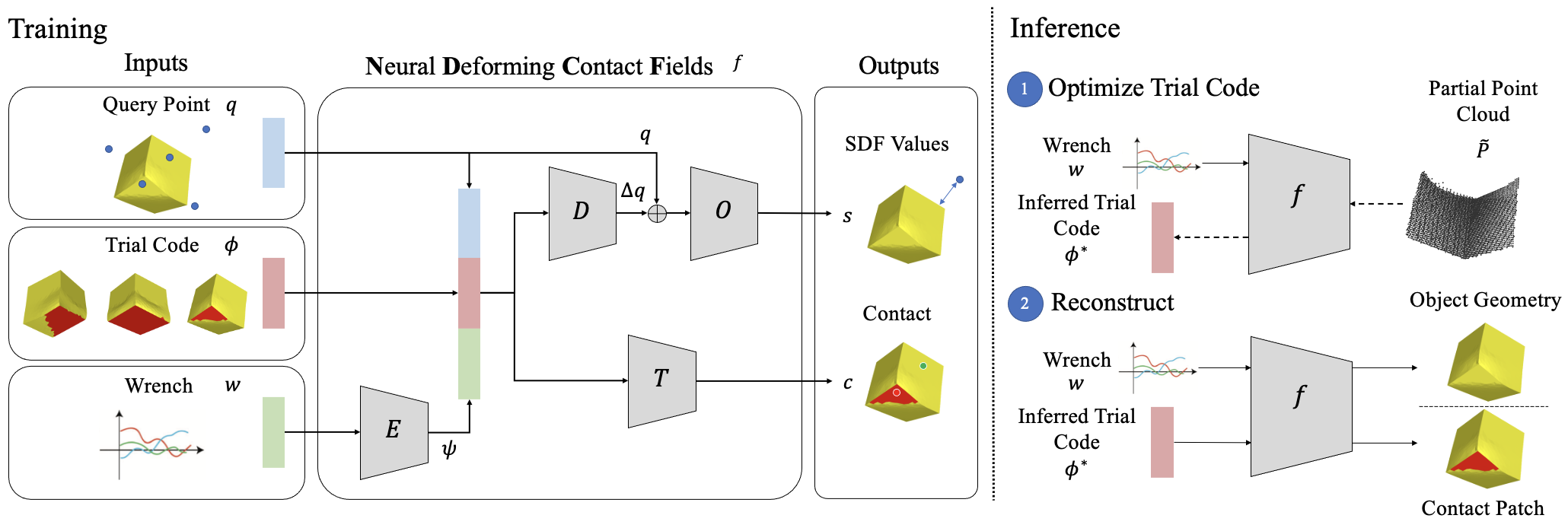}
    \caption{Method Overview: We propose \textbf{Neural Deforming Contact Fields}, to jointly reason over deformable object geometries and contact patches, conditioned on wrench feedback. We represent both geometry and contact as an implicit method, predicting an SDF $s$ and a contact probability $c$ for query points $q$. At inference time, we take in a partial point cloud and wrench value and recover a desired trial latent code, which can be used to reconstruct the full geometry and contact patch.}
    \label{fig:method_overview}
\end{figure*}

Our work builds on the recent advances in neural implicit and signed-distance field (SDF) representations of object geometry that have recently gained significant attention in computer vision \cite{park2019deepsdf,sitzmann2020implicit,rebain2022attention} and robotics \cite{driess2022learning,wi2022virdo,wi2022virdo++}. These representations offer several advantages over traditional geometric object and environment models. One of their key benefits is the ability to handle objects and environments with complex, non-linear shapes, such as deformable objects and cluttered scenes. Unlike traditional geometric models, which often rely on a fixed set of vertices and edges, neural implicit representations and SDFs encode the shape and behavior of objects as continuous functions. Recent work has utilized neural implicit representations to effectively model deformable object geometries~\cite{wi2022virdo, wi2022virdo++}. However, existing methods have largely ignored geometric representations of contact. Our approach brings contact geometry to center stage, considers object deformations, and allows for far greater flexibility than classical models such as a point contacts~\cite{manuelli2016localizing}, line contact~\cite{van2022learning,kim2021active,ma2021extrinsic}, or planar patch contact \cite{lynch1996stable,yu2016more,zhou2016convex} that are typically limited to rigid-body objects and have limited representation capacity.



\section{Related Work}

\textbf{3D geometry representation for deformable objects:}
Several robotics manipulation studies adopt discrete geometry representations including meshes \cite{ficuciello2018fem,sengupta2019tracking, sundaresan2022diffcloud} and GNNs \cite{ma2021learning, shi2022robocraft, lin2022learning}. These approaches often require identification of connectivities \cite{lin2022learning, sundaresan2022diffcloud} and special node-tracking techniques during inference \cite{sengupta2019tracking}. Though these methods seem promising for cloth folding \cite{lin2022learning, huang2022mesh, sundaresan2022diffcloud} and dough manipulation \cite{shi2022robocraft}, it is unclear how to use these representations for complex volumetric objects, predicting dense contacts, and integration with force transmissions.  On the other hand, prior works on dense 3D geometry representations provide high-fidelity surface reconstructions for complex volumetric geometries \cite{ park2019deepsdf, DIF, sitzmann2020implicit}. These approaches include  SDF \cite{wi2022virdo, wi2022virdo++,Palafox2021ICCV, mu2021sdf, ortiz2022isdf}, occupancy~\cite{Niemeyer2019ICCV}, and volume density \cite{park2021nerfies, driess2022learning, driess2022reinforcement}, which directly use raw vision output (RGB-D) without annotations.  
Unlike existing work modeling deformable object geometries which either assume known contact geometries~\cite{wi2022virdo} or do not explicitly reason over contact~\cite{wi2022virdo++}, our proposed work directly incorporates contact into the representation.

\textbf{Contact representations:} Prior works on recovering contact locations have utilized a set of points ~\cite{manuelli2016localizing} and lines ~\cite{kim2021active, ma2021extrinsic, van2022learning} as representations for contacts. These representations are appropriate for rigid or relatively non-deformable object interactions. However, the sponge, being a highly deformable tool, often undergoes contacts that occur in the form of patches, making the conventional representations insufficient.

Contact patches have been utilized as a representation when using collocated sensing at the point of contact~\cite{li2013control, sutanto2019learning, suresh2022midastouch}. We are interested instead in \textit{extrinsic} contact interactions, where we have little to no sensing at the location of contact.

Most related to our work by way of representing contact is the Neural Contact Field proposed by~\citet{higuera2022neural}. Similar to our method, the Neural Contact Field treats contact as an implicit function which maps from a query point to a contact probability. However, they assume rigid, known object models and do not consider multi-modal inputs, focusing on tactile sensing. In contrast, we consider the case of deformable objects, where the change in geometry due to contact is unknown, and focus on a joint representation of deformable object geometry and contact, learned from visuo-tactile inputs.

\section{Method}

\subsection{Formulation} \label{sec:ndcf_formulation}

We propose Neural Deforming Contact Fields (NDCF), a learned multimodal implicit representation that simultaneously reasons over grasped object deformations, wrench feedback, and contact patches. The key insight of our method is to jointly predict the geometry of the object and the contacts on the object surface, rather than assume contacts are provided~\cite{wi2022virdo} or reasoning about contacts indirectly~\cite{wi2022virdo++}. We choose to represent the contact patch implicitly as a neural field, which allows for contact patches of varying shapes and topologies to be represented with a single model. Additionally, by learning the deformed object geometry and contact patch jointly, we can enforce physical priors during training and inference, ensuring that contacts lie on the surface of the deformable object.

Our method is designed to incorporate feedback from common robotic sensors. As such, we assume access to a partial point cloud $\tilde{P}\in \mathbb{R}^{N\times 3}$ of the segmented object and wrench feedback at the robot wrist $\vec{w}\in \mathbb{R}^6$. Then, given a query point $\vec{q}\in \mathbb{R}^3$, we predict its SDF value $s\in \mathbb{R}$ and the likelihood that it is in contact $c \in [0,1]$:
\begin{align*}
    (s, c) = f(\tilde{P}, \vec{w}, \vec{q})
\end{align*}
The deformable object surface is given by the zero-level set of the SDF value:
\begin{align} \label{eq:infer_surface}
    S = \{ \vec{q} \; | \; (s=0, c) = f(\tilde{P},\vec{w}, \vec{q}) \}
\end{align}
The zero level set can be recovered through Marching Cubes \cite{lorensen1987marching} or ray tracing methods and can easily be converted to a point cloud or a geometric mesh $\mathcal{M}$ \cite{cignoni2008meshlab}.

The contact patch is given by the intersection of the object surface with points classified as in contact, where $\epsilon$ is the binary classification threshold:
\begin{align} \label{eq:infer_contact_patch}
    C = S \cap \{\vec{q} \;|\; (s, c> \epsilon) =  f(\tilde{P},\vec{w}, \vec{q}) \}
\end{align}

\subsection{Architecture}

We propose a neural network architecture for learning NDCFs from data. The full architecture can be seen in Fig.~\ref{fig:method_overview}.

\vspace{3pt}
\noindent\textbf{1. Network Inputs:} We train the network to reconstruct geometry and contact patches from static interactions between a deforming object and its environment. To this end, we adopt the encoder-less geometric reasoning popular for learning implicit geometries~\cite{mu2021sdf,park2019deepsdf,wi2022virdo,wi2022virdo++}. By training without an encoder, we additionally decouple the input point cloud distribution from the network training. We introduce a trial code $\vec{\phi}\in \mathbb{R}^{L}$ that captures the current deformation and contact occurring in the scene. The latent space of $\vec{\phi}$ is learned simultaneous to the training of the network weights.

The wrench $\vec{w}$ is encoded through a neural network $E$ into a latent vector $\vec{\psi} = E(\vec{w})\in \mathbb{R}^{L}$ used to introduce force reasoning into the network. The latent vectors $\vec{\phi}$ and $\vec{\psi}$ are jointly input to later parts of the network and trained together, allowing for joint reasoning over forces, deformations, and contact patches.

\vspace{3pt}
\noindent\textbf{2. Signed Distance Field:} We follow~\citet{wi2022virdo,wi2022virdo++} in representing deforming object SDFs by decomposing into a nominal SDF and deformation field, relating the deformed geometry to the nominal geometry. First, we represent the nominal SDF implicitly, with a neural network $O$, taking in a query point $\vec{q}\in \mathbb{R}^3$ and predicting a SDF $s\in \mathbb{R}$ as $s = O(\vec{q})$.

The second component is a predicted deformation field. This network predicts how each point in space can be deformed \textit{back} to the nominal geometry, represented implicitly with the network $D$. As the deformation is dependent on the particular interaction, we provide this network with our latent vectors $(\vec{\phi}, \; \vec{\psi})$ to inform the deformation. $D$ then maps from a query point $\vec{q} \in \mathbb{R}^3$ to a deformation $\Delta \vec{q} \in \mathbb{R}^3$ as $\Delta q = D(\vec{q} | \vec{\phi}, \vec{\psi})$. Following~\citet{wi2022virdo,wi2022virdo++}, the final SDF prediction becomes:
\begin{align*}
    s = O(\vec{q} + D(\vec{q} | \vec{\phi}, \vec{\psi}))
\end{align*}

In this paper, we learn an \acro~for a single object. In future work, we plan to extend this framework to learning across multiple classes of object by adding an additional object latent code~\cite{wi2022virdo,mu2021sdf,park2019deepsdf}. 

\noindent\textbf{3. Contact Field:} We predict the likelihood of contact at every point in space. To make the prediction we use a neural network $T$. Similar to deformation, our contact is dependent on the particular interaction, so we provide this network with our latent vectors $(\vec{\phi}, \vec{\psi})$. $T$ then maps from a query point $\vec{q}\in \mathbb{R}^3$ to a contact probability $c\in [0,1]$:
\begin{equation*}
    c = T(\vec{q} | \vec{\phi}, \vec{\psi})
\end{equation*}

\subsection{Training} \label{sec:method-training}
Our training is composed of two steps: first, we pretrain the object module $O$ such that the nominal geometry (without contact) is accurately represented by this module. Next, we train the entire architecture end-to-end to model deformations and contact patches.

\vspace{3pt}
\noindent\textbf{1. Pretraining the Object Module:} We pretrain our object module $O$ to fit the SDF of our nominal object geometry. Given a dataset of sampled points around the nominal geometry, $\Omega=\{ \vec{q}_i, s_i^*, \vec{n}_i^* \}_i$, we train with the following loss: 
\begin{align*} \label{eq:l_sdf}
\begin{split}
    \mathcal{L}_{sdf} = \frac{1}{|\Omega|} \sum_{i=1}^{|\Omega|} |O(\vec{q}_i) - s_i^*|  + \xi \frac{1}{|\Omega_S|} \sum_{i=1}^{|\Omega_S|} (1 - \langle \nabla O(\vec{q}_i), \vec{n}_i^* \rangle)
\end{split}
\end{align*}
where $\Omega_S$ is the subset of sampled surface points and $\xi$ weights the SDF and normal losses. The normal loss encourages the predicted and ground truth normals to align on the object surface. The predicted surface normal can be recovered by differentiating the SDF output with respect to the query point: $\nabla O(\vec{q}_i) = \partial O(\vec{q}_i)/\partial \vec{q}_i$. The gradient can be calculated using backpropagation. The weights of $O$ are frozen after pretraining. 

\vspace{3pt}
\noindent\textbf{2. Loss Formulation:} Our model is trained on a dataset of interactions provided as: 
\begin{align*}
\mathcal{D} = \{ (\Omega_1, \vec{w}_1, \vec{\phi}_1), (\Omega_2, \vec{w}_2, \vec{\phi}_2), \dots, (\Omega_N, \vec{w}_N, \vec{\phi}_N) \}    
\end{align*}
For each interaction we have a set of sampled points $\Omega_i = \{\vec{q}_j, s_j^*, \vec{n}^*_j, c^*_j\}_i$, as well as the generated trial code $\vec{\phi}_i$ and wrench for the example $\vec{w}_i$. For each example $(\Omega_i, \vec{w}_i, \vec{\phi}_i)$, we calculate the training loss given by:
\begin{align*}
\begin{split}
    \mathcal{L}_{train} =& \mathcal{L}_{sdf} + \alpha \mathcal{L}_{embedding} + \beta \mathcal{L}_{deform} + \omega \mathcal{L}_{chamfer} \\ &+ \gamma \mathcal{L}_{contact}
\end{split}
\end{align*}
where $\mathcal{L}_{sdf}$ encourages the final predicted SDF values and normals to match the ground truth, and is defined as:
\begin{align*} 
\begin{split}
    \mathcal{L}_{sdf} =& \frac{1}{|\Omega_i|} \sum_{j=1}^{|\Omega_i|} |O(\vec{q}_j + D(\vec{q}_j| \vec{\phi}_i, \vec{\psi}_i)) - s_j^*|  \\
    & + \xi \frac{1}{|\Omega_{i,S}|} \sum_{j=1}^{|\Omega_{i,S}|} (1 - \langle \nabla O(\vec{q}_j + D(\vec{q}_j|\vec{\phi}_i)), \vec{\psi}_i), \vec{n}_j^* \rangle)
\end{split}
\end{align*}
where $\Omega_{i,S}$ are the points in $\Omega_i$ lying on the surface of the object, $\xi$ weights the SDF and normal losses, and $\nabla O(\vec{q}_j + D(\vec{q}_j|\phi_i, E(\vec{w}_i))=\frac{\partial O(\vec{q}_j + D(\vec{q}_j|\vec{\phi}_i, \vec{\psi}_i))}{\partial \vec{q}_j}$ is computed using backpropagation.

$\mathcal{L}_{embedding} = ||\vec{\phi}_i||_2^2$ is used to ensure the learned latent space is well-formed~\cite{park2019deepsdf}. $\mathcal{L}_{deform} = \frac{1}{|\Omega_i|} \sum_{j=1}^{|\Omega_i|} || D(\vec{q}_j| \vec{\phi}_i, \vec{\psi}_i) ||_2^2$ is used to embed the prior that smaller deformations are preferred to complex large deformations. 

The loss $\mathcal{L}_{chamfer}$ is a Chamfer distance used to encourage the predicted nominal surface, derived by adding the query points that lie on the surface to their predicted deformations, matches the ground truth nominal surface. Let
$$P = \{ \vec{q}_j + D(\vec{q}_j | \vec{\phi}_i, \vec{\psi}_i) | s_j^*=0 \}$$ be the predicted nominal surface and $P^*$ be the ground truth nominal surface point cloud. Then,

\begin{equation*}
    \mathcal{L}_{deform} = \text{CD}(P, P^*)
\end{equation*}
where CD is the Chamfer distance between two point clouds:
\begin{align} 
\begin{split} \label{eq:chamfer}
    \text{CD}(P_1, P_2) = &\frac{1}{|P_1|} \sum_{\vec{x}\in P_1} \min_{\vec{y} \in P_2} ||\vec{x} - \vec{y}||_2^2 \\ &+ \frac{1}{|P_2|} \sum_{\vec{x}\in P_2} \min_{\vec{y} \in P_1} ||\vec{x} - \vec{y}||_2^2
\end{split}
\end{align}

Finally, we supervise our contact field using $\mathcal{L}_{contact}$ defined as follows:
\begin{equation}
    \mathcal{L}_{contact} = \frac{1}{|\Omega_{i,S}|} \sum_{j=1}^{|\Omega_{i,S}|} \text{BCE}(T(\vec{q}_j| \vec{\phi}_i, \vec{\psi}_i), c^*_j)
\end{equation}
where BCE is the Binary Cross Entropy loss. Notably, we only evaluate binary classification on the \textit{surface}, that is the set of points $\Omega_{i,S}$. This avoids having to learn a 2D contact surface directly, rather inheriting the geometric surface predicted by the SDF, while maintaining the flexibility of the implicit function in terms of shape and topology.

Training is achieved by solving the following optimization:
\begin{equation}
    \vec{\theta}^*, \vec{\phi}_i^* = \underset{\vec{\theta}, \vec{\phi}}{\mathrm{argmin}} \sum_{i=1}^{|\mathcal{D}|} \mathcal{L}_{train}(\mathcal{D}_i)
\end{equation}
As in other encoder-less methods, we simultaneously train our \textit{trial codes} $\vec{\phi}_i$ alongside the weights of our network.

\subsection{Inference}

We assume access to a partial point cloud $\tilde{P}$ and wrench reading $w$. We then can perform inference to find the trial code $\phi^*$ for the example using the following optimization:
\begin{equation} \label{eq:inference}
    \phi^* = \min_{\phi} \frac{1}{|\tilde{P}|}\sum_{i=1}^{|\tilde{P}|} O(\tilde{P}_i + D(\tilde{P}_i | \vec{\phi}, E(\vec{w}))) + \eta ||\vec{\phi}||_2^2
\end{equation}
This optimization finds the latent vector $\vec{\phi}$ that places all the partial points on the surface of the generated geometry, additionally conditioned on the wrench $\vec{w}$, thus matching the generated NDCF to the observations. The second loss term is used to regularize the prediction~\cite{park2019deepsdf} and prevents drifting away from well formed latent vectors. The geometry and contact patch can then be recovered from the partial point cloud and wrench measurements, as detailed in Sec.~\ref{sec:ndcf_formulation}.

\section{Implementation}

\subsection{NDCF Implementation}\label{sec:ndcf_implementation}

Our wrench encoder $E$ is implemented as multi-layer perceptron (MLP) with a single hidden layer with size $16$. Recent studies in neural implicit methods suggest the best method for conditioning outputs is using Hyper Networks, where conditioning vectors are used to predict the weights of an additional network, which takes in the query point and predicts the field value~\cite{wi2022virdo,wi2022virdo++,DIF,rebain2022attention}. As such we implement our deformation module $D$ and contact module $T$ as hypernetworks:
\begin{align*}
    \Delta\vec{q} &= D(\vec{q} | H_D(\vec{\phi}, \vec{\psi})) \\
    c &= T(\vec{q} | H_T(\vec{\phi}, \vec{\psi}))
\end{align*}
Here $H_D$ and $H_T$ predict the weights of each module MLP. We add additional regularization terms to regularize the predicted weights.
\begin{align*}
    \mathcal{L}_{hyper} =& \zeta (\frac{1}{|H_D(\vec{\phi},\vec{\psi})|}||H_D(\vec{\phi},\vec{\psi})||_2^2 \\
    &+ \frac{1}{|H_T(\vec{\phi},\vec{\psi})|}||H_T(\vec{\phi},\vec{\psi})||_2^2)
\end{align*}
$\zeta$ is a weighting on the loss. The predicted deformation MLP has a single hidden layer of size 256 while the predicted contact MLP has two hidden layers of size 256. The object nominal SDF module $O$ is implemented as a MLP with 2 hidden layers of size 256. All models are implemented using Pytorch. We set latent space dimension of $\vec{\phi}$ and $\vec{\psi}$ to be in $L=16$.

\subsection{Training and Inference Details}

NDCF Pretraining is performed using the Adam Optimizer with learning rate $1e-5$. We set normal loss weighting term $\xi=0.01$. The pretraining is run for 50000 epochs to effectively memorize the nominal object geometry.

NDCF Training is performed using the Adam Optimizer with learning rate $1e-4$. The latent codes $\phi_i$ are initialized from the zero mean Gaussian with standard deviation $0.1$. From Sec.~\ref{sec:method-training}, we set $\alpha=1e-3,\beta=1.0,\omega=0.01,\gamma=0.1,\xi=0.01$. From Sec.~\ref{sec:ndcf_implementation}, we set $\zeta=1e-6$. We train our method for 200 epochs.

NDCF Inference is performed with the Adam Optimizer with learning rate $2e-3$. Latent codes, as during training, are initialized from a zero mean Gaussian with standard deviation $0.1$. In each experiment, we use a validation set to select the choice of $\eta$ in Eq.~\ref{eq:inference}, the contact probability threshold $\epsilon$, and the number of gradient descent iterations to perform. In particular, we perform a grid search and choose inference hyper-parameters with the best performance. We report the selected hyper-parameters for each experiment in Sec.~\ref{sec:experiments}.

\subsection{Data Collection} \label{sec:data_collection}

\subsubsection{Simulation} \label{sec:data_sim}
The training methodology described in Sec.~\ref{sec:method-training} relies upon having strong supervision of sampled points and their corresponding SDF and contact patches. In this work, we utilize Isaac Gym~\cite{makoviychuk2021isaac}, a GPU-based physics simulator, to collect simulated deformable-object interactions. Isaac Gym implements 3D Finite Element Method (FEM) on the GPU and has been experimentally validated for accuracy in rigid-deformable interactions~\cite{narang2021sim}.

\begin{figure}
    \centering
    \includegraphics[width=0.4\textwidth]{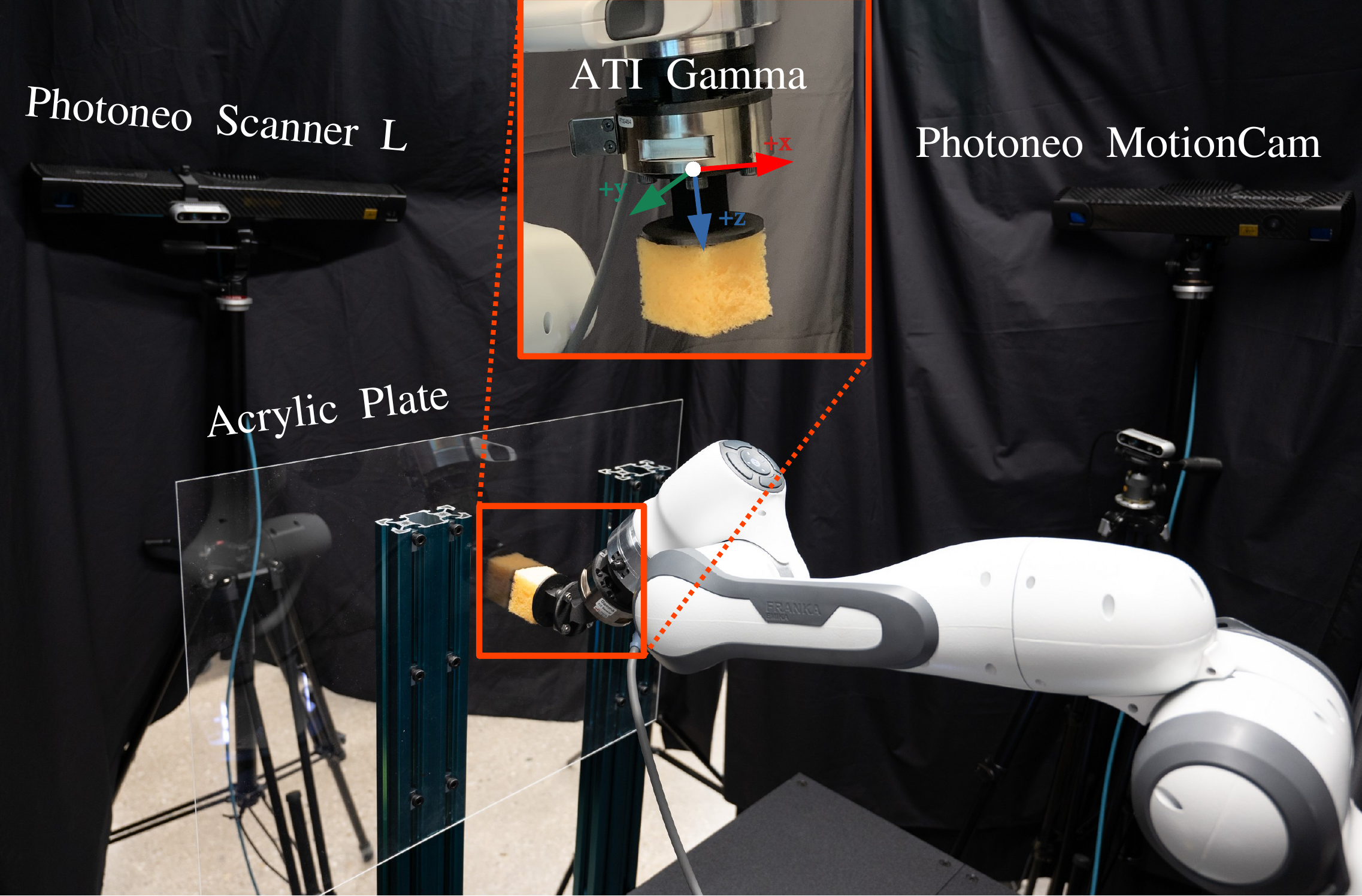}
    \caption{Real-world experiment setup with 1 Pandas arm, 2 depth cameras (Photoneo ScannerL and Photoneo MotionCam-3D color), 1 force-torque sensor (ATI Gamma), and 1 46mm sponge mounted on the force-torque sensor. ScannerL was used for labelling while MotionCam was used for the partial point cloud inputs.}
    \label{fig:experiment_setup}
\end{figure}

\begin{figure}
    \centering
    \includegraphics[width=0.45\textwidth]{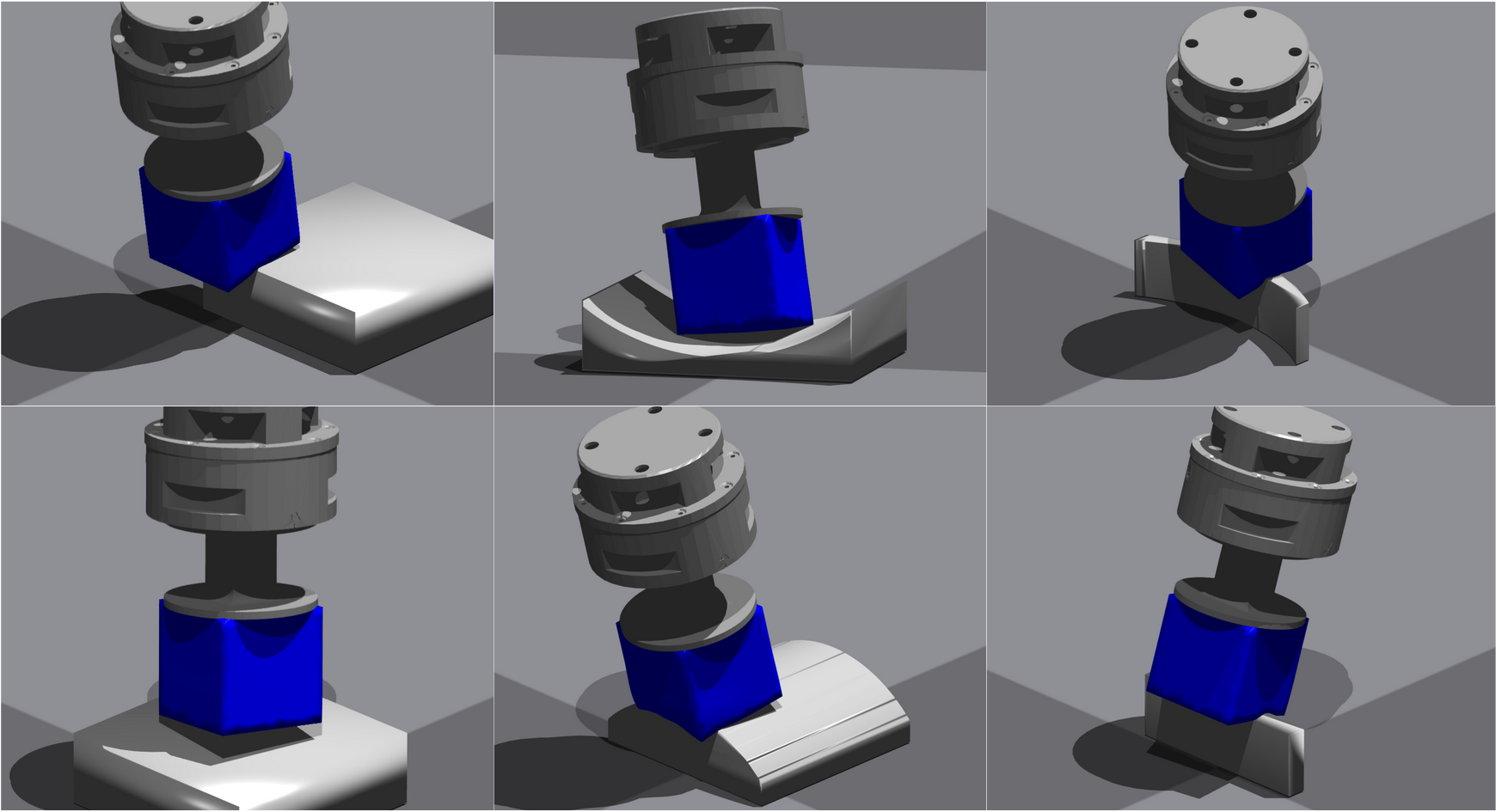}
    \caption{Visualization of generated environments and sampled interactions for our training dataset in Isaac Gym. Environments per column are \textbf{box}, \textbf{curves}, and \textbf{ridges} from left to right.}
    \label{fig:random_terrain}
\end{figure}

\begin{figure}
    \centering
    \includegraphics[width=0.4\textwidth]{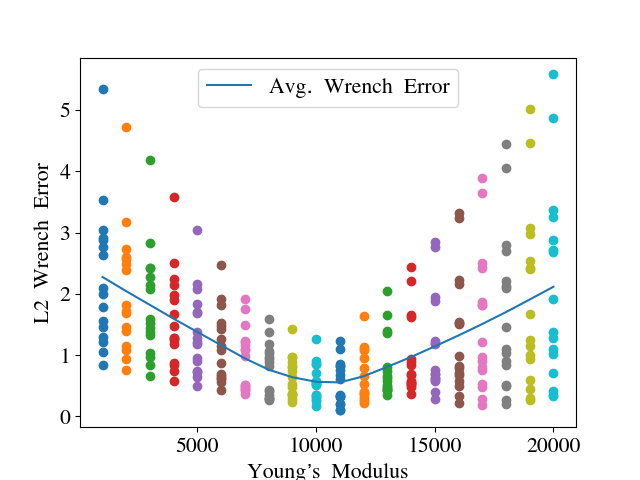}
    \caption{We perform system identification to match simulated material properties to our real object setup. 16 real setup presses are recreated in simulation and the simulated wrench is compared to that registered in the real world. Points indicate individual simulation trial comparisons to the real world data, colored for visual clarity by the used Young's Modulus. We found the optimal Young's Modulus to be $E=1.1e4$ Pa.}
    \label{fig:sysid_wrench_error}
\end{figure}

In this work, our object is a 46mm cube sponge rigidly attached to a Franka Emika Panda end effector, with an ATI Gamma F/T sensor mounted between (see Fig.~\ref{fig:experiment_setup}). We recreate the sponge in Isaac Gym, rigidly attached to the wrist mounting geometry used in the real world (see Fig.~\ref{fig:random_terrain}). Isaac Gym FEM assumes homogenous material properties and requires specification of the Poisson's ratio $\nu$ and Young's Modulus $E$. We set $\nu=0.1$ (a low value) to reflect the fact that because the sponge is a porous material with air gaps, it has a significantly smaller transverse elongation due to compression than a solid object. To determine $E$ we perform system identification. On the real system, we collect a series of 16 presses onto a flat surface and record the wrench response measured by the F/T sensor. We then recreate those presses in simulation, by moving the simulated wrist to the same pose, and measure the simulated response. We perform a line search over Young's Modulus values, using the euclidean distance between real and simulated wrenches to evaluate match. We find that $E=1.1e4$ Pa best matched our robot setup in simulation (see Fig.~\ref{fig:sysid_wrench_error}).

We collect a dataset of simulated presses to train our NDCF. We use three scenarios: first is a box, second is a randomly generated curved surface, and third is a randomly generated ridge terrain (see Fig.~\ref{fig:random_terrain}). In each scenario, we sample random orientations for the wrist and randomize the starting position of the gripper to enable interactions near the edges of objects as well as on the surfaces, then lower straight down until a randomly sampled distance between 3mm and 1cm past the point of initial contact. The resulting data is saved and processed to form dataset $\mathcal{D}$, including ground truth geometries, feedback wrenches, and contact labels.

In total, 3000 environments and interactions are sampled, 1000 per each environment type. For each interaction, we sample 40000 query points off surface and 20000 query points on the surface and generate ground truth SDF values, contact patches, and normal labels. We augment our dataset by rotating all inputs around the tool z-frame (frame shown in Fig.~\ref{fig:experiment_setup}), as our system exhibits symmetry due to the shape of the tool.

To evaluate our method, we also collect an additional dataset of new interactions with full labels. We generate 300 new press interactions, 100 per environment, and prune examples where the resulting contact area was less than $1\times 10^{-6}mm^2$, for an evaluation set of 298 interactions. To provide partial geometric information, we place eight cameras spaced evenly around the wrist facing the tool to capture partial point clouds of the sponge in contact from multiple angles, which are used during inference, along with the simulated wrench feedback. Examples of partial views are shown in Fig.~\ref{fig:sim_results_geometry}.

\subsubsection{Physical Robot} \label{sec:data_physical}
We also demonstrate our method's effectiveness in the real-world using a physical robot. In order to provide useful evaluation of our method, we design a real-world test bed from which we can derive contact patch labels. However, we emphasize that these labels are not used for training and are only for evaluation purposes. Shown in Fig.~\ref{fig:experiment_setup}, our real world setup places a clear acrylic plate in front of the Franka Emika Panda with the sponge mounted rigidly to the robot end effector. A Photoneo Phoxi 3D scanner (L) is placed opposite the acrylic. When the robot moves the sponge into contact with the acrylic, the Photoneo depth scan sees through the acrylic and locates the contacting face of the sponge. By calibrating the position of the acrylic, we can estimate the contact patch from the resulting depth scan by thresholding the points near the face of the acrylic. Examples of resulting labels can be found in Fig.~\ref{fig:real_results_flat}. We use these as the ground truth contact patch to evaluate our methods on the real world data.

We use the wrench feedback from the mounted ATI gamma as the wrench input. To get our partial point cloud, a Photoneo MotionCam-3D Color (M+) is mounted next to the robot viewing the contact interaction. The sponge is segmented from the resulting point cloud and provided as the partial geometry to each method.

We collect 48 different interactions with the acrylic sheet. We randomize the orientation of the sponge before pressing by sampling an offset angle in the range $[-0.3, 0.3]$ for rotations about the $(x,y,z)$ axes relative to the nominal pose, which is the robot ATI Gamma frame (shown in Fig.~\ref{fig:experiment_setup}) with the $z$ axis pointing directly at the plate and the $x$ axis pointing normal to the table underneath. The robot is moved to an offset position then pressed straight forward towards the acrylic 1cm past the point of initial contact.

\section{Experiments} \label{sec:experiments}

\subsection{Baseline}

We compare our implicit representation method to an explicit counterpart. Point clouds are flexible shape representations; however, they lack connectivity and are fundamentally discrete which limits their resolution and ability to capture high-frequency details. Here, we implement an explicit baseline representation that directly predicts both the geometry and contact patch as \textit{point clouds}. We adopt the point cloud encoder-decoder architecture of \cite{xie2020grnet} and condition the bottleneck feature on both the partial pointcloud $\tilde{P}$ and wrench $\vec{w}$. We implement two point cloud decoders, one to output deformed object geometries and the other to output the contact patch predictions. 

\begin{equation*}
    P, C = f(\tilde{P}, \vec{w})
\end{equation*}

We use the following loss to train:
\begin{equation*}
    \mathcal{L}_{baseline} = \xi_1 \text{CD}(P, P^*) + \xi_2 \text{CD}(C, C^*) + \xi_3 \text{CD}_{uni}(C, P)
\end{equation*}
where $P$ is the estimated object surface point cloud and $P^*$ is the ground truth surface point cloud. Similarly, $C$ is the estimated contact patch and $C^*$ is the ground truth contact patch. The final loss term incorporates the knowledge that we know that the contact patch should lie on the surface of the object. This can be encouraged in the learned  model by a \textit{uni-directional} Chamfer distance, which only calculates the distance from the contact patch $C$ to the nearest points on the geometry $P$.
\begin{equation*}
    \text{CD}_{uni}(C, P) = \frac{1}{|C|} \sum_{\vec{x}\in C} \min_{\vec{y} \in P} || \vec{x} - \vec{y} ||_2^2
\end{equation*}
We used Adam Optimizer with learning rate 1e-4, set $\xi_1 = \xi_2 = \xi_3 = 1e4$, and ran 200 epochs for training on the same dataset used to train the NDCF. Upon reconstructing geometry from the point cloud $P$, we apply a meshing algorithm to convert the output to a mesh~\cite{curless1996volumetric}. 

The use of an encoder-decoder structure means the baseline is sensitive to the distribution of the partial view $\tilde{P}$ seen for training examples, e.g., the placement of the cameras. As such, we perform augmentation during training, selecting subsets of the camera views so that the encoder is somewhat robust to the available partial view. We note our decoder-only method is invariant to the input distribution, since partial views are not used during training and rather are used in the loss during inference. As such, we do not need to perform any augmentation when training our method, in comparison to the baseline.


\begin{figure}
    \centering
    \includegraphics[width=\linewidth]{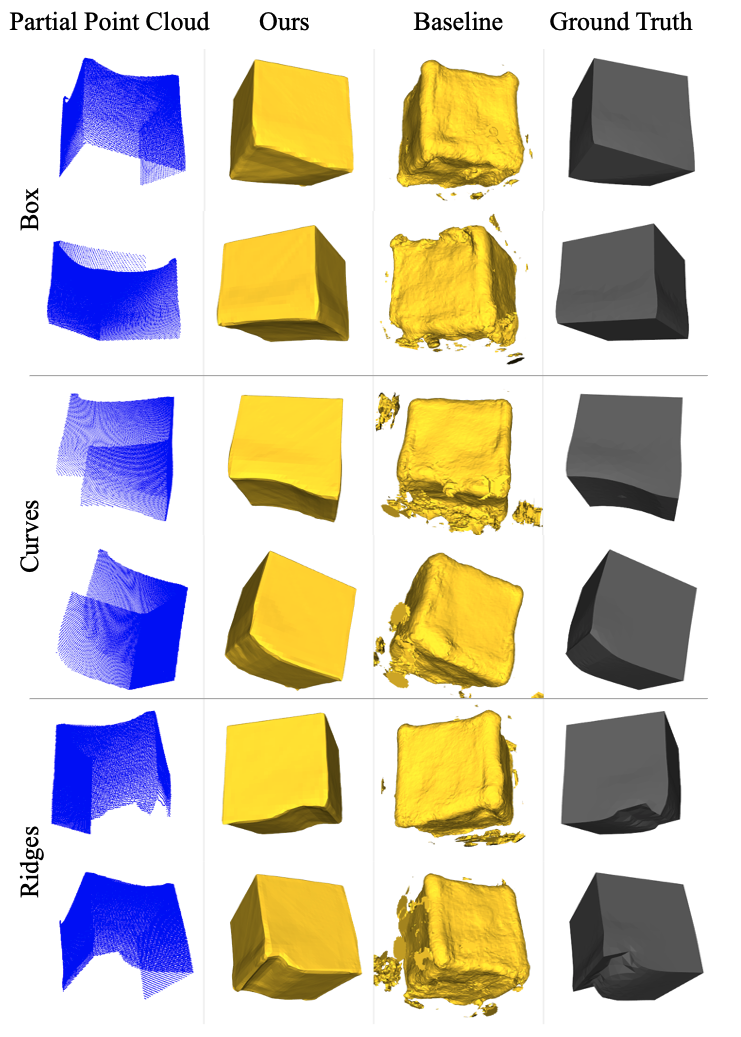}
    \caption{Selected simulation geometry results from each of our test environments (Box, Curves, and Ridges). Our proposed method NDCF shows crisp reconstructions with high retention of detail, while the baseline exhibits artifacts and lacks geometric details.}
    \label{fig:sim_results_geometry}
\end{figure}

\begin{figure}
    \centering
    \includegraphics[width=0.85\linewidth]{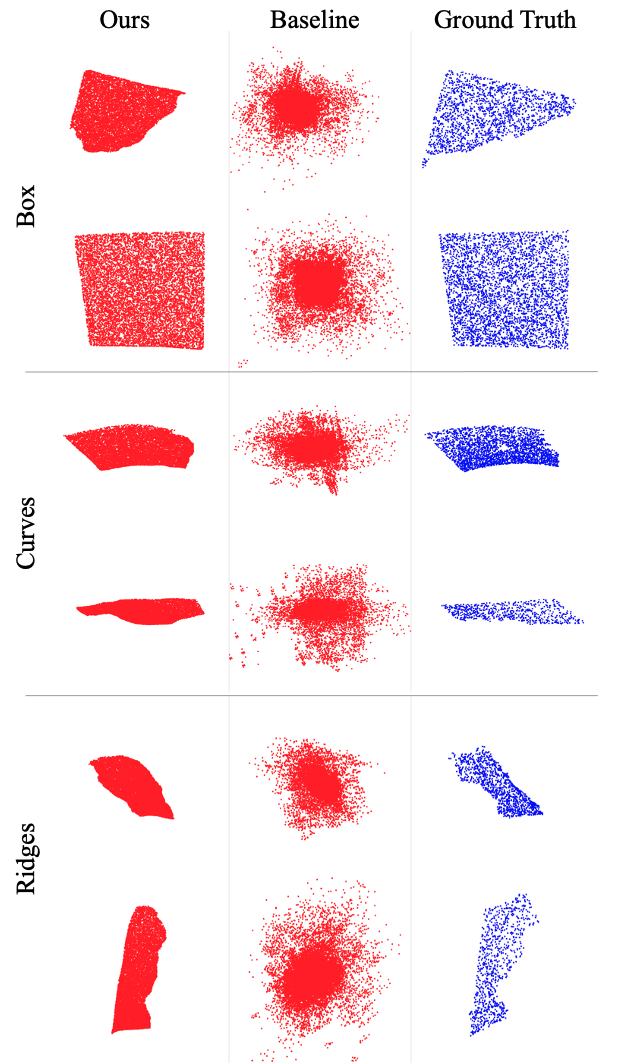}
    \caption{Simulation contact patch estimation results for examples from Fig.~\ref{fig:sim_results_geometry}. Our proposed method NDCF captures the variety in contact shapes while the baseline estimates exhibit high variance with points predicted far from the contact patch.}
    \label{fig:sim_results_contact_patches}
\end{figure}

\subsection{Metrics}

\noindent \textbf{1. Geometry:} To evaluate the deformed object geometry, we use two metrics. The first metric is the Chamfer Distance (see Eq.~\ref{eq:chamfer}) of the predicted surface point cloud to the ground truth surface point cloud. For NDCF, we generate the surface point cloud $P$ by sampling from the generated object mesh. For the point cloud baseline, we directly compare the generated object surface point cloud to the ground truth. To make for fair comparison, we down-sample each resulting point cloud and ground truth point cloud to 10,000 points.

The second metric is a Volumetric IoU of the predicted mesh $\mathcal{M}$ and ground truth mesh $\mathcal{M}^*$. The Volumetric IoU is defined as the quotient of the volume of the intersection of the meshes by the union of the meshes,
\begin{equation*}
    IoU(\mathcal{M}, \mathcal{M}^*) = \frac{|\mathcal{M} \cap \mathcal{M^*}|}{|\mathcal{M} \cup \mathcal{M}^*|}
\end{equation*}
We estimate the volumes by sampling 100,000 points near the object and determining which points lie inside the predicted and ground truth meshes~\cite{mescheder2019occupancy}.

\noindent \textbf{2. Contact Patch:} We evaluate the contact patch geometries using the Chamfer Distance (see Eq.~\ref{eq:chamfer}). For NDCF, we use rejection sampling on the surface to recover the contact patch, evaluating the contact probability at each sample and accepting those above $\epsilon$. For the point cloud baseline, we directly compare the generated contact patch point cloud to the ground truth. To make for fair comparison, we downsample the predicted and ground truth contact patch point clouds to 300 points. We found that the baseline did not spread the points evenly through the predicted patch, so we use a voxel-downsampling technique to better sample from the point cloud volume fairly.

\subsection{Simulated Experiments} \label{sec:sim_exp}


{\renewcommand{\arraystretch}{1.3}
\begin{table*}[t]
\centering
\caption{Model Performance on Simulated and Real-World Data}\label{tab:sim_evaluation}
\begin{tabular}{l c c c c c c} 
\toprule
    & \multicolumn{2}{c}{Deformed Geometry (sim)} & Contact Patch (sim) & Inference Time (sim) & Meshing Time (sim) & Contact Patch (real-world) 
    \\
    \cline{2-7}
    & $\text{CD}_{mm^2}$  & $IoU$ &  $\text{CD}_{mm^2}$ & Seconds & Seconds & $\text{CD}_{mm^2}$ \\
    \hline
    Ours &  \textbf{0.910 (0.158)} & \textbf{0.987 (0.007)} & \textbf{22.840 (40.414)} & 0.173 (0.006) & 1.906 (0.378) & \textbf{39.502 (23.797)} \\
    Baseline~\cite{xie2020grnet} &  5.746 (1.091) & 0.817 (0.058) & 36.537 (19.046) & \textbf{0.026 (0.001)} & -- & 57.853 (39.502)\\
\bottomrule
\end{tabular}
\end{table*}
}



We first evaluate our method on 298 unseen simulated interactions, as detailed in Sec.~\ref{sec:data_sim}. We use a validation set of 90 press interactions, 30 from each environment, to select inference hyper-parameters. We select $\eta=0.001$, $\epsilon=0.2$, and run inference for $100$ gradient steps.

Our method successfully predicted a contact patch in all but two cases, in which it failed to predict any points on the deformed surface with contact probability above $\epsilon=0.2$. To avoid skewing baseline performance, we remove these two examples when calculating performance metrics for both methods.

Full performance results are shown in Table~\ref{tab:sim_evaluation}. We see that our method outperforms the proposed baseline on all three metrics, reflecting better prediction of both object geometry and contact patches. We also show the runtime of each method - while the inference procedure does take longer than a forward pass of a model, we still are able to run our method at roughly 0.5Hz, including the time to perform meshing via Marching Cubes. Further tuning of inference parameters could trade off performance and runtime for specific applications. Additional simulation performance analysis can be found in Appendix~\ref{sec:app_cd_details}.


In Fig.~\ref{fig:sim_results_geometry} and Fig.~\ref{fig:sim_results_contact_patches}, we show the predicted geometry and contact patch for several examples from each environment. Our method, NDCF, shows high retention of geometric details, such as the curvature of the mesh in the first ``Curves'' example. The proposed baseline is much more noisy, with rounded edges and artifacts appearing off the surface from spuriously generated points. Our method also shows the ability to capture diverse contact shapes accurately, whereas the baseline method, while often clustered near the patch, is noisy, with contact points predicted at times far from the object surface. Our method, can directly take advantage of the geometric reasoning of NDCF, and as such the contact patches lie cleanly on the predicted surface of the object.

\subsection{Physical Robot Experiments} \label{sec:real_res}

\begin{figure}
    \centering
    \includegraphics[width=\linewidth]{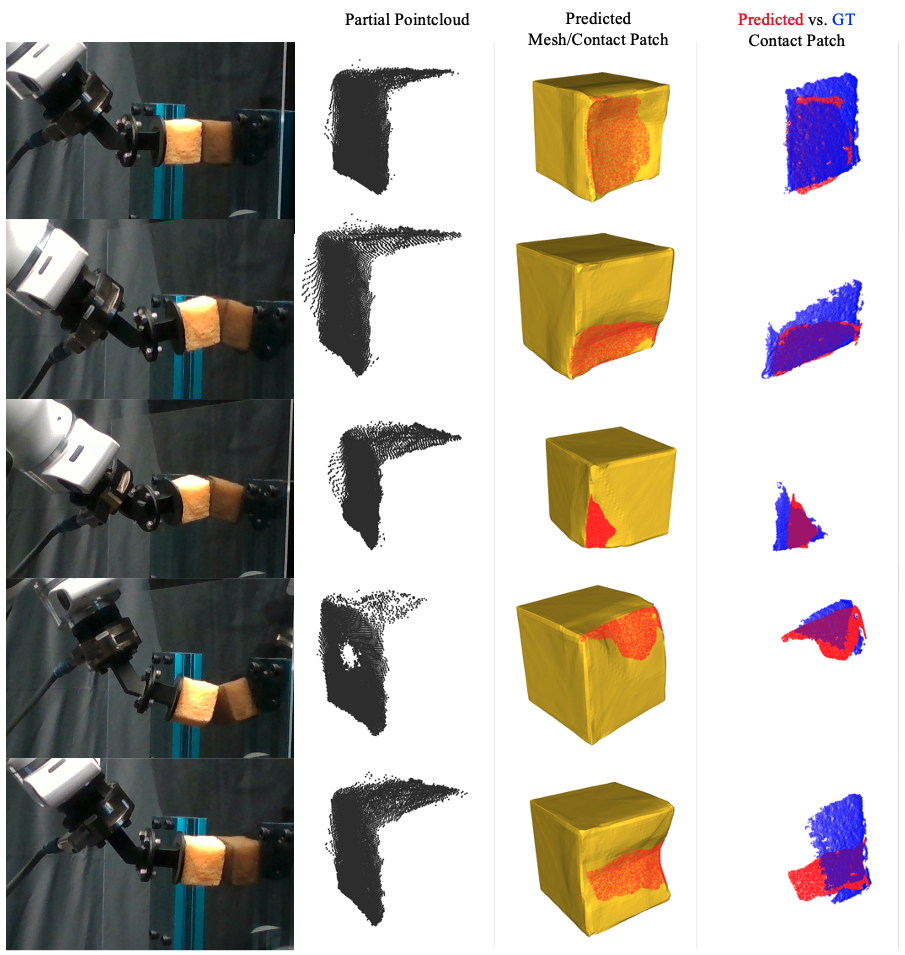}
    \caption{NDCF geometry and contact patch estimation on real world interactions with a flat surface. Our results show that NDCF trained on simulated data transferred to real world examples. Predicted meshes accurately predict where deformation occurs for each interaction and the contact patches are accurate to the measured ground truth patch. Note that the right three columns are rotated to view the contacting face of the object. The last row shows a failure case, where the contact occurs opposite the camera and thus is harder to detect.}
    \label{fig:real_results_flat}
\end{figure}

\begin{figure}
    \centering
    \includegraphics[width=0.7\linewidth]{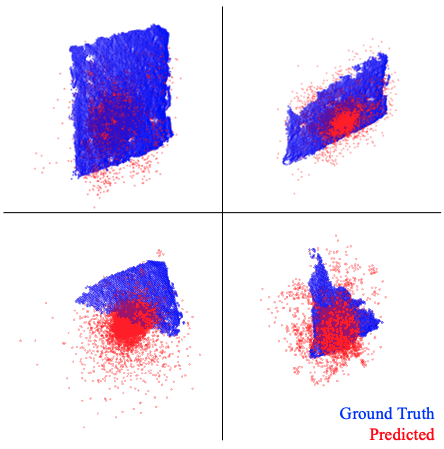}
    \caption{Clockwise from top-left: baseline contact predictions for the first four examples in Fig.~\ref{fig:real_results_flat}. While the main cluster of points is located near the ground truth patch, the baseline predictions exhibit variance, with points predicted far from the patch and object surface.}
    \label{fig:real_results_baseline}
\end{figure}

Here, we evaluate our methods ability to generalize from the simulated environment to the real world one. We use 48 physical robot interactions with ground truth contact patch labels, collected as described in Sec.~\ref{sec:data_physical}. We use a set of 16 validation interactions, collected in the same manner, to choose inference hyper-parameters. We select $\eta=0.1$, $\epsilon=0.2$, and run inference for $100$ gradient steps. A higher $\eta$ is intuitive, since the real world sensing is more noisy than the simulated data, thus a term preventing the latent from drifting out of distribution prevents over-fitting to sensor noise.

Fig.~\ref{fig:real_results_flat} shows our methods performance on the real world data for several interactions with varying contact patch shapes. In each case, the predicted mesh appears to closely match the interaction and the contact patches are good estimates of the ground truth. Fig.~\ref{fig:real_results_baseline} shows baseline predictions for the first four examples of Fig.~\ref{fig:real_results_flat}. Similar to the simulation results, we found that the baseline often clustered points roughly near the patch but exhibited noise and often predicts points far from the contact.

Over the 48 test interactions, we evaluate the quality of the contact patch using the Chamfer Distance metric. As we do not have ground truth geometries, we forgo deformed geometry evaluation. As shown in Tab.~\ref{tab:sim_evaluation}, our method outperforms the baseline on patch CD.  

\subsection{YCB Interaction Experiments}

\begin{figure}
    \centering
    \includegraphics[width=\linewidth]{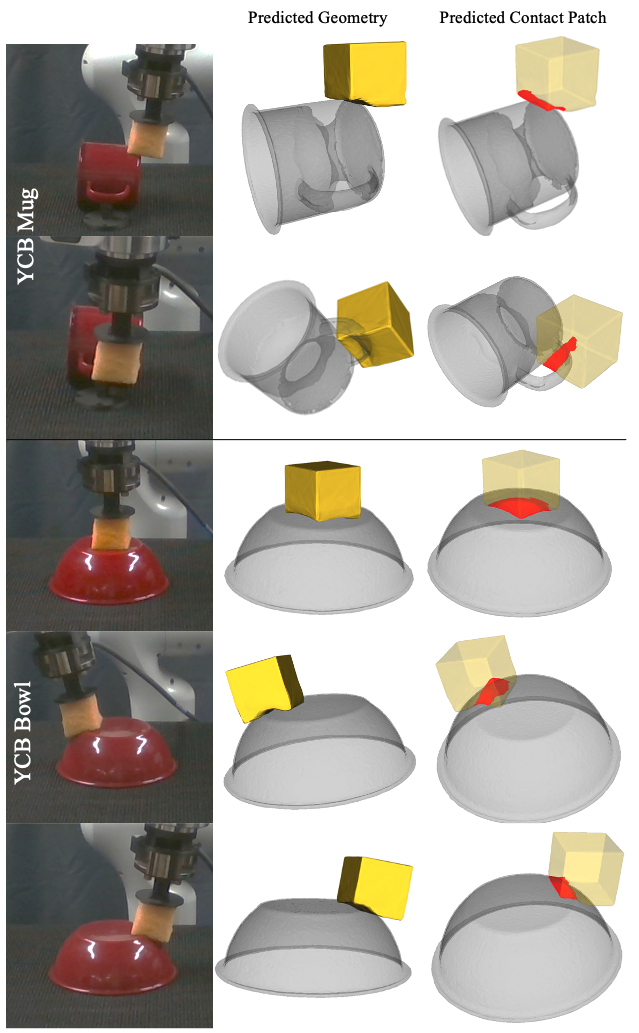}
    \caption{Qualitative results of applying NDCF to poke interactions with the YCB mug and bowl object. We find that for interactions with these objects, our method is able to recover feasible geometric completions and contact patches that are close to the ground truth object surface.}
    \label{fig:poke_results}
\end{figure}

As a final experiment, we qualitatively investigate our method's performance on objects in the real world. We place the mug and bowl object from the YCB dataset~\cite{calli2017yale} in front of the robot. Poke interactions are prerecorded by a demonstrator, then repeated autonomously. We use the same robot and sensor setup described in Sec.~\ref{sec:data_physical}, but now use partial point clouds combined from both the Photoneo MotionCam and Photoneo Scanner L as the partial visual data. We found using the inference hyper-parameters from Sec.~\ref{sec:sim_exp} worked well for this data.

We show the qualitative results for three presses on each object in Fig.~\ref{fig:intro_figure} and Fig.~\ref{fig:poke_results}. The ``ground truth'' pose of the object used in visualization is approximated by performing ICP on a scan of the object from the Photoneo sensors with the robot moved clear. We see that NDCF is able to complete reasonable geometries and contact patches, including differentiating between flat contacts and contacts near the edge of objects, as well recognizing the ``ridge'' like contact of the mug handle.

To provide an approximate quantitative measure of real world performance, we measure the average squared distance from the predicted contact patch points to the surface of the object mesh (this can be seen as a uni-directional Chamfer Distance). Our method achieves 15.379 mm² for the Mug and 2.728 mm² for the Bowl.

\section{Conclusion} 
\label{sec:conclusion}

We present Neural Deforming Contact Fields (NDCFs), a representation that jointly reasons over extrinsic contact patches, deformable object geometry, and force transmitted via contact. Our method represents both the object geometry and contact patch using neural implicit fields, which makes the representation flexible enough to handle complex contact shapes and object deformations. 

Our results indicate that we can utilize NDCFs to recover deformed object geometries and corresponding contact patches with high accuracy, and outperforms an explicit representation baseline method which uses point cloud representations. Additionally, we showed that we could transfer the representation to the real world without finetuning and demonstrated the ability to recover geometries and contact patches given real interaction data.

So far, our demonstration of NDCF has been limited to a single tool interacting statically with an environment. To handle multiple objects, we can condition our nominal object model using an object latent code to distinguish geometries, and introduce a materials latent code to address the case of identical geometry with varying material properties. For dynamic interactions, our representation can be augmented with dynamics in the latent space to predict future trial and wrench latent codes, conditioned on actions.

Another limitation of this current work is that it does not directly consider the surrounding environment. Here, we primarily focused on demonstrating shared reasoning over geometries and contact, independent of the environment geometry. However, in certain cases (see last row Fig.~\ref{fig:real_results_flat}) the partial geometry of the deformable object and the reaction wrench may not fully disambiguate the underlying deformations and contacts, as many varying contact configurations can yield similar observations. 



\bibliographystyle{plainnat}
\bibliography{references}

\begin{appendices}

\section{Input Ablations}

To better elaborate on the role of the wrench input, we show model performance on our simulated test data as increasing amounts of noise are injected into the wrench input. Noise is added by sampling zero-mean noise with variance scaled by the size of the reading. That is, for our wrench input $\vec{w}$ we add a percentage $p$ noise to each term as follows:

\begin{equation*}
    \vec{w}_i = \vec{w}_i + \epsilon; \epsilon \sim \mathcal{N}(0.0, p\cdot \vec{w}_i)
\end{equation*}

We investigate how increasing the percentage of the scale as the standard deviation of the noise affects model performance. Table~\ref{tab:sim_noise} shows model performance as we increase the injected noise. Patch \% indicates the percentage of examples for which a patch is correctly predicted (it is possible for no points on the surface to have contact likelihood above $\epsilon$). We see that for mild noise (10\%), the performance stays largely the same. At lower noise levels, information from the point cloud may be enough to correct for the noise in the wrench. At higher noise levels, the estimation of both geometry and contact deteriorates, both in terms of percent of examples where patches are predicted and the quality of the patch predictions, indicating the importance of the wrench value.

{\renewcommand{\arraystretch}{1.3}
\begin{table}
    \centering
    \begin{tabular}{l c c c}
    \toprule
       & \multicolumn{1}{c}{Deformed Geometry (sim)} & \multicolumn{2}{c}{Contact Patch (sim)}  \\
       \cline{2-4}
       Noise (\%) & $\text{CD}_{mm^2}$  & Patch \% & $\text{CD}_{mm^2}$ \\
       \hline
       0 & 0.910 (0.158) & 99.3 & 22.840 (40.414) \\
       10 & 0.958 (0.233) & 97 & 26.644 (41.411) \\
       50 & 2.895 (12.144) & 95.7 & 128.865 (292.644) \\
       100 & 15.304 (77.485) & 83.3 & 434.111 (774.389) \\
    \bottomrule
    \end{tabular}
    \caption{Model Performance with Noisy Wrench Inputs}
    \label{tab:sim_noise}
\end{table}
}

\section{Contact Patch Performance Details} \label{sec:app_cd_details}

\begin{figure}
    \centering
    \includegraphics[width=\linewidth]{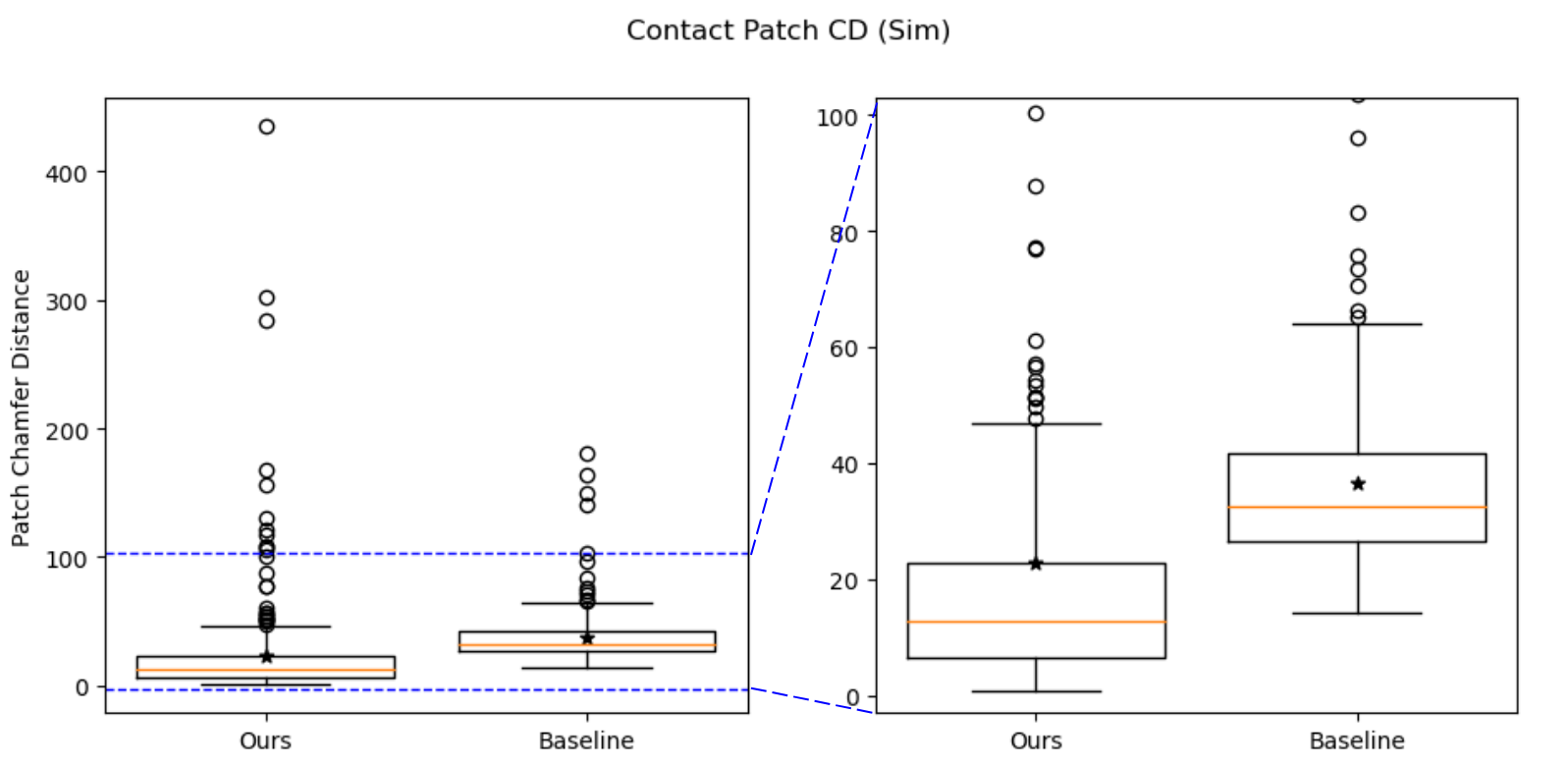}
    \caption{Boxplot of Simulated Test Contact Patch CD. Stars indicate the mean of each method. The right pane shows a zoomed view to highlight performance details. Our method outperforms the baseline on mean, median, and quartile performance, but does have outliers with high error. }
    \label{fig:patch_cd_boxplot}
\end{figure}

\begin{figure}
    \centering
    \includegraphics[width=\linewidth]{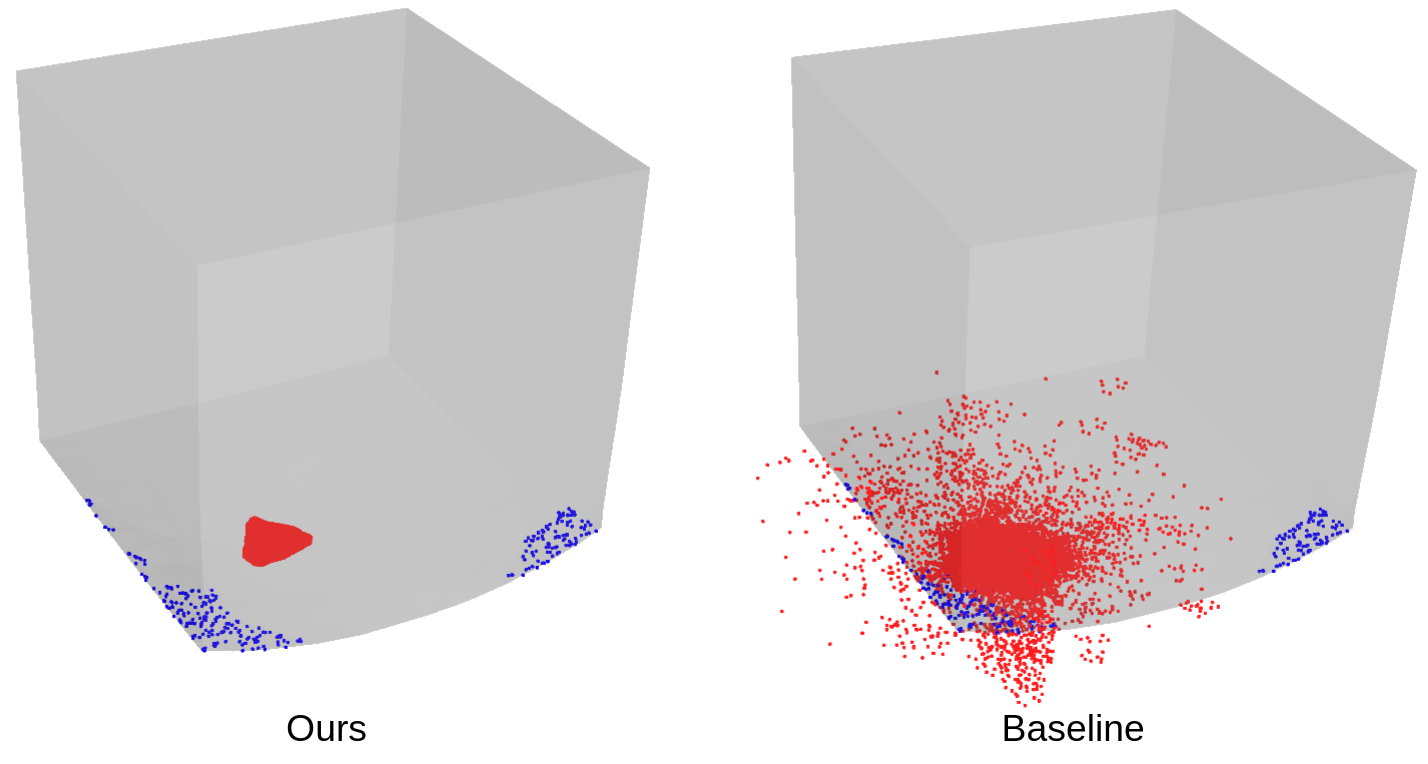}
    \caption{Contact Patch \textcolor{red}{predictions} (in red) vs. \textcolor{blue}{ground truth} (in blue) for our method and the baseline on our method's outlier result in the simulation dataset. We see that our method predicts a patch that could have similar composite wrench feedback and similar geometry.}
    \label{fig:patch_outlier}
\end{figure}

We provide further details on the simulated contact patch Chamfer Distance (CD) performance. In Fig.~\ref{fig:patch_cd_boxplot} we show the spread of the contact patch CD as a boxplot. Our method shows better median and quartile performance than the baseline. Our method also outperforms on average, but we see that our average is pulled up by outliers. This also explains the higher standard deviation in Tab.~\ref{tab:sim_evaluation}.

In Fig.~\ref{fig:patch_outlier}, we show our method and the baseline on our method's outlier example from Fig.~\ref{fig:patch_cd_boxplot}. This is a difficult example due to its small deformation and contact, as well as the split contact, which can alias other contact configurations. We see that our method predicts a patch that could have a similar wrench feedback to the composite wrench of the actual interaction. The baseline prediction has a cluster of points that is similarly located, but the baseline's tendency to predict a noisy spread of points means the evaluation is much more kind to the baseline, with a CD of $103.5mm^2$, than to our method, with a CD of $435.9mm^2$, despite the fact that neither method predicted accurately on this example (note, as explained in Sec.~\ref{sec:experiments}, each point cloud is sampled to 300 points before evaluating CD). This result helps contextualize why our method has higher error on a small number of outlier examples. 

We found that for most examples, however, as shown qualitatively in Fig.~\ref{fig:sim_results_contact_patches} and in the distribution of CD values in Fig.~\ref{fig:patch_cd_boxplot}, our method performs favorably to the baseline. 

\end{appendices}


\end{document}